 \documentclass[smallabstract,smallcaptions]{dccpaper}

\usepackage{epsfig}
\usepackage{citesort}
\usepackage{amsmath}
\usepackage{amssymb}
\usepackage{color}
\usepackage{url}

\newlength{\figurewidth}
\newlength{\smallfigurewidth}

\usepackage{graphicx}
\usepackage{booktabs}
\usepackage{multirow}
\usepackage{makecell}
\usepackage[table, svgnames, dvipsnames]{xcolor}

\newcommand{\bmmc}[1]{\bm{\mathcal{#1}}}

\newcommand{\bm}[1]{\boldsymbol{\mathcal{#1}}}
\DeclareMathOperator*{\argmax}{arg\,max}

\makeatletter
  \newcommand\figcaption{\def\@captype{figure}\caption}
  \newcommand\tabcaption{\def\@captype{table}\caption}
\makeatother

\begin{document}

\title
{\large
\textbf{Transferable Learned Image Compression-Resistant Adversarial Perturbations}
}

\author{%
Yang Sui$^{1}$, Zhuohang Li$^{2}$, Ding Ding$^{3}$, Xiang Pan$^{3}$, \\
Xiaozhong Xu$^{3}$, Shan Liu$^{3}$, Zhenzhong Chen$^{4}$ \\[0.5em]
{\small\begin{minipage}{\linewidth}\begin{center}
\begin{tabular}{ccccccc}
$^{1}$Rutgers University $^{2}$Vanderbilt University $^{3}$Tencent America $^{4}$Wuhan University\\
\url{yang.sui@rutgers.edu} 
\end{tabular}
\end{center}\end{minipage}}
\thanks{The work of Yang Sui was done during the internship at Tencent America.}
\thanks{The work of Zhuohang Li was done during the visit at Tencent America.} 
\thanks{The work of Zhenzhong Chen was done during the visit at Tencent.}
}

\maketitle
\thispagestyle{empty}

\begin{abstract}
    Adversarial attacks can readily disrupt the image classification system, revealing the vulnerability of DNN-based recognition tasks. While existing adversarial perturbations are primarily applied to uncompressed images or compressed images by the traditional image compression method, i.e., JPEG, limited studies have investigated the robustness of models for image classification in the context of DNN-based image compression. With the rapid evolution of advanced image compression, DNN-based learned image compression has emerged as the promising approach for transmitting images in many security-critical applications, such as cloud-based face recognition and autonomous driving, due to its superior performance over traditional compression. Therefore, there is a pressing need to fully investigate the robustness of a classification system post-processed by learned image compression. To bridge this research gap, we explore the adversarial attack on a new pipeline that targets image classification models that utilize learned image compressors as pre-processing modules. Furthermore, to enhance the transferability of perturbations across various quality levels and architectures of learned image compression models, we introduce a saliency score-based sampling method to enable the fast generation of transferable perturbation. Extensive experiments with popular attack methods demonstrate the enhanced transferability of our proposed method when attacking images that have been post-processed with different learned image compression models.
\end{abstract}

\section{Introduction}

Deep Neural Network(DNN)-based models are known to be vulnerable against adversarial examples~\cite{szegedy2014intriguing}, which are carefully perturbed images that are unsuspicious to human eyes but can cause deep learning models to produce incorrect or malicious predictions. Although this phenomenon was initially discovered on image classification models~\cite{szegedy2014intriguing}, research on adversarial examples has since then quickly spread to many critical domains in computer vision, such as facial recognition~\cite{dong2019efficient}, and object detection~\cite{xie2017adversarial}. Initial studies~\cite{szegedy2014intriguing,xie2017adversarial} typically assume a white-box access to the target model. Later, black-box attacks~\cite{ilyas2018black} are also developed where the details of the target model are unknown to the adversary.

Existing works predominantly study adversarial examples in the context of uncompressed images. However, when deployed to real-world systems, including cloud-based image analysis services and edge image recognition systems (e.g., object recognition for autonomous driving and face recognition for security services), in order to save communication bandwidth or computation cost, typically images are first compressed using some compression algorithms before being fed into the classification system to get predictions. In light of this scenario, a branch of studies~\cite{shin2017jpeg,wang2020towards,zhou2018transferable} has investigated JPEG-compression-resistant adversarial attacks.


Recently, Learned Image Compression (LIC) has rapidly become the primary method for transmitting images under bit-rate constraints due to its superior performance. Specifically, LIC frameworks \cite{balle2018variational,cheng2020learned,sui2024corner} have recently evolved, showing substantial rate-distortion performance improvement over standard image compression \cite{bross2021overview} such as JPEG\cite{wallace1991jpeg}, JPEG2000\cite{taubman2002jpeg2000}, and BPG\cite{bellard2016bpg}, owing to the remarkable representation ability of DNNs. The fundamental structure of LIC is the auto-encoder framework with entropy minimization constraints, which employs the DNN-based encoder and decoder for image compression and reconstruction. 

LIC has been quickly adopted in many application scenarios as a pre-processing module for various image classification tasks and has also been proposed in ISO/IEC JTC 1/SC29/WG1 M93073 in the 93rd JPEG-AI Meeting. Despite its wide deployment, the robustness of the LIC-powered image recognition pipeline remains under-explored. To fill this research gap, in this paper, we aim to investigate the robustness of \textbf{\textit{Learned Image Compression Classification System (LICCS)}} by launching specialized adversarial attacks that are optimized for this pipeline. Further, in the LICCS, to-be-classified compressed images have unknown compression quality levels determined by LIC. Therefore, exploring the robustness across all quality levels presents a unique challenge. Overall, the goal of our investigation is to answer the following research questions:

\begin{itemize}
\vspace{-2mm}
\item \textit{How robust are the LICCSs against adversarial perturbations?}
\vspace{-2mm}
\item \textit{How transferable are adversarial perturbations across \textbf{various quality levels} under the same/different LIC model architectures?}
\vspace{-2mm}
\end{itemize}

\begin{figure*}[t]
\centering
  \includegraphics[width=0.34 \linewidth]{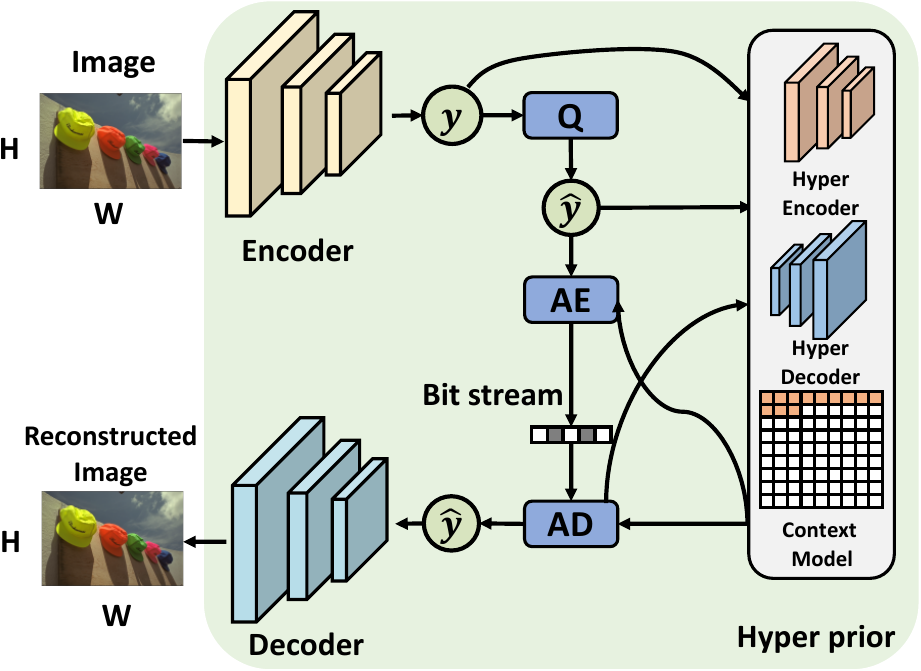}
  \includegraphics[width=0.62 \linewidth]{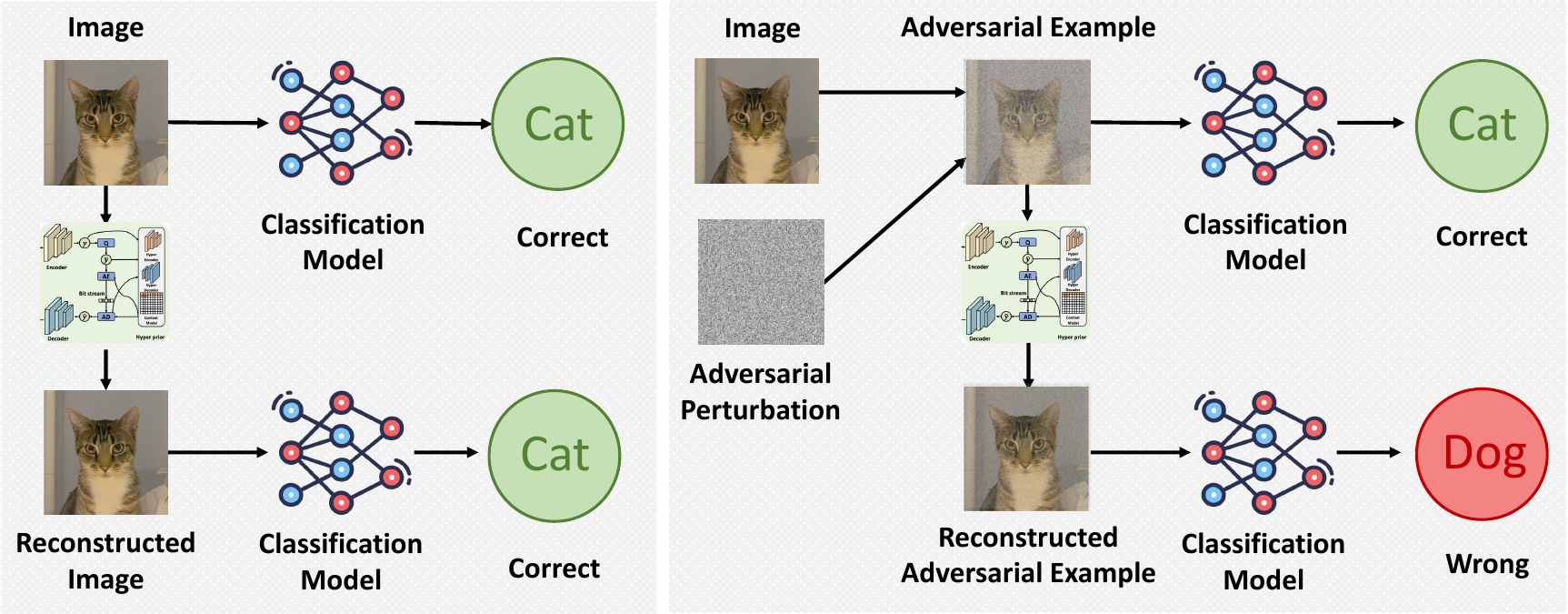}

\caption{Our proposed adversarial attack pipeline against the LICCS. Left: The framework of the LIC (described in Section \ref{method:LIC}); Middle: LICCS pipeline. Both the original image and the reconstructed image are classified with the correct label ``Cat"; Right: After the adversarial attacks, although the adversarial examples are classified with the correct label, its reconstructed image through the LIC is misrecognized as the wrong label ``Dog".}
\label{fig:pipeline}
\vspace{-2mm}
\end{figure*}

To address these questions, we conduct a series of empirical investigations on a typical LICCS pipeline, which is illustrated in Fig. \ref{fig:pipeline}.
To evaluate the robustness of the LICCS, we first consider a white-box scenario where the attacker has the ability to access the details of models.
Next, to evaluate how the attack can be generalized in the black-box scenario, we measure the transferability of perturbations across various unknown quality levels under the same LIC model architecture.
We find through our experiments that the LICCS framework is naturally, to some extent, resilient to transferable attacks, as the attack performance shows significant disparities across different quality levels.
To improve the transferability across the quality levels, we propose a saliency score-based sampling method that performs ensemble attacks on the most influential quality levels of LIC models, which show the highest adversarial impact on the LICCS across all quality levels.
Specifically, we measure the collective coverage of affected curves of all combinations of surrogate models to calculate the saliency score. Based on these scores, we select the highest influential combination consisting of top-$K$ quality levels, which are further utilized to launch attacks to generalize to all quality levels. Our contributions can be summarized as follows: (1) We investigate an adversarial attack pipeline for the LICCS, utilizing LICs as pre-processing modules for the image classification model. To the best of our knowledge, we are the first work to investigate the robustness of LICCS. (2) To measure the robustness of LICCS and the transferability of its adversarial perturbations, we conduct a series of experiments in white-box and black-box scenarios. Based on the black-box results, we observed that the neighboring quality levels are more significantly affected when attacking a certain quality level. (3) To improve the transferability of the attack across different quality levels and architectures, we propose a saliency score-based sampling method that enables generating transferable perturbations with limited model access. (4) To improve the transferability of the attack across different quality levels and architectures, we propose a saliency score-based sampling method that enables generating transferable perturbations with limited model access.




\section{Transferable Attack against LICCS}
\label{sec:liccs}

\subsection{Learned Image Compression}
\label{method:LIC}

As illustrated in Fig. \ref{fig:pipeline} (left), the input image is converted into a latent representation via a non-linear transformation. The latents are then encoded by an arithmetic encoder to produce the bit stream for storage. To reconstruct the image, the bit stream is decoded by an arithmetic decoder and fed into the main DNN-based decoder to generate a reconstructed image. Specifically, suppose $g_a(\cdot)$, $g_s(\cdot)$ are the non-linear transforms. Let $\bmmc{X}$ and $\bmmc{\hat{X}}$ denote the original input and reconstructed images, respectively. Let $\bmmc{Y}$ and $\bmmc{\hat{Y}}$ denote the pre-quantized and quantized latent representation, respectively, then the deep learning-based image compression can be described as:
\begin{equation}
\vspace{-2mm}
\begin{aligned}
\bmmc{Y} &= g_a (\bmmc{X}),\\
\quad \bmmc{\hat{Y}} &= \text{AD}(\text{AE}(Q(\bmmc{Y}))),\\
\quad \bmmc{\hat{X}} &= g_s(\bmmc{\hat{Y}}),\\
\end{aligned}
\label{eqn:lic}
\end{equation}
where $Q(\cdot)$ is the quantization process. $\text{AE}$ and $\text{AD}$ denote the arithmetic encoding and decoding processes, respectively. $\bmmc{\hat{X}}$ represents the reconstructed image. To reduce the bit rate, a hyper-prior is used as side information to estimate the mean and scale parameters of latents predicted from the entropy model, including a hyper encoder and hyper decoder. Furthermore, an auto-regressive context model is integrated into the hyper-prior framework to boost the R-D performance. Since we aim to explore the LICCS, which prioritizes the reconstructed image over the bit-rate. Therefore, we omit the hyper-prior and context model for simplicity.

\subsection{LICCS Attack}

Fig. \ref{fig:pipeline} (right) illustrates the attack pipeline for generating adversarial examples in conjunction with or without the LIC. Let $\bmmc{X}$, $y$ denote the image and ground-truth label. Given a classification model $f(\cdot)_{i}$ predicts the probability of the image belonging to class $i$, adversarial attacks aim to generate an adversarial perturbation $\bmmc{\delta}$ embedded on $\bmmc{X}$ so that the new adversarial image can misclassify the classification model $f(\cdot)_{i}$. It can be formulated as:
\begin{equation}
\vspace{-2mm}
\begin{aligned}
\argmax \limits_{i} f(\bmmc{X}+\bmmc{\delta})_{i} \neq y, \quad \textrm{s.t.} \enspace \| \bmmc{\delta} \|_{p} \leq \epsilon, \\
\end{aligned}
\label{eqn:attack}
\end{equation}
where $\epsilon$ denotes the perturbation budget to ensure the induced distortion is imperceptible. 

Unlike prior works that focus on the adversarial attack toward reconstruction image quality of LIC \cite{sui2023reconstruction,chen2023towards}, in this paper, we primarily investigate the robustness of a LIC-based classification system. To perform an adversarial attack on an image within the LICCS as illustrated in Fig. \ref{fig:pipeline} (right), the goal is to introduce the adversarial perturbation $\bmmc{\delta}$ to the source image $\bmmc{X}$ that causes the reconstructed adversarial examples $g_s(Q(g_a(\bmmc{X}+\bmmc{\delta})))$ to be misclassified by the classification model. Then, Eq. \ref{eqn:attack} is extended as follows:
\begin{equation}
\vspace{-4mm}
\begin{aligned}
\argmax \limits_{i} f(g_s(Q(g_a(\bmmc{X}+\bmmc{\delta}))))_{i} \neq y, \quad \textrm{s.t.} \enspace \| \bmmc{\delta} \|_{p} \leq \epsilon. \\
\end{aligned}
\label{eqn:multi_lic_attack}
\end{equation}


\subsection{White-box Attacks Evaluation}

To evaluate the robustness of the LICCS, we first perform the white-box attacks by solving Eq. \ref{eqn:multi_lic_attack}. The detailed results of Top-1 accuracy of LICCS (\texttt{cheng2020} \cite{cheng2020learned} LIC model and ResNet-20 classification model) attacked by the PGD attack~\cite{madrytowards} on the CIFAR-10 dataset is shown in Table \ref{table:whiteboxresults} of Section \ref{sec:exp}. Given a pre-trained classification model with 91.25 $\%$ top-1 accuracy on the CIFAR-10 dataset in LICCS, when generating adversarial examples with $\epsilon=16$, the accuracy of LICCS decreases to 18.75$\%$ and 5.53$\%$ in quality levels 1 and 6, demonstrating considerable vulnerability to attacks. Hence, we conclude \textbf{the LICCS is vulnerable to white-box attacks}, mainly contributing to the inherent DNN-based classification model.

\subsection{Transferability from Black-box Attacks}

\begin{figure*}[t]
    \begin{minipage}[b]{0.45\linewidth}

    \setlength{\tabcolsep}{2pt}
        \begin{center}
            \tabcaption {Top-1 accuracy of PGD black-box attack results of \texttt{cheng2020} \cite{cheng2020learned} model. Each row/column corresponds to a surrogate/target model with a given quality level.}
    	\scalebox{0.7}{
    		\begin{tabular}{cccccccc}
    			\toprule
                   Quality & 1 & 2 & 3 & 4 & 5 & 6 \\ \hline
    			\multicolumn{7}{c}{$\epsilon=4$, \quad $\alpha=1$, \quad $iters=10$} \\ 
    			\midrule
        1 & 35.59\% & 57.54\% & 68.61\% & 79.25\% & 84.50\% & 87.19\% \\ \hline
        2 & 43.71\% & 37.86\% & 58.07\% & 76.14\% & 83.01\% & 85.65\% \\ \hline
        3 & 45.85\% & 46.15\% & 35.82\% & 69.49\% & 79.02\% & 83.00\% \\ \hline
        4 & 48.72\% & 56.92\% & 58.16\% & 37.28\% & 53.77\% & 66.30\% \\ \hline
        5 & 49.22\% & 59.32\% & 62.38\% & 47.77\% & 34.83\% & 41.01\% \\ \hline
        6 & 49.55\% & 60.20\% & 65.07\% & 57.97\% & 39.10\% & 32.53\% \\ \hline
    			\bottomrule
    		\end{tabular}
    		}
    	 \label{table:blackboxresults}
    \end{center}
    \end{minipage}
    \hfill
    \begin{minipage}[b]{0.48\linewidth}
        \includegraphics[width=0.8\linewidth]{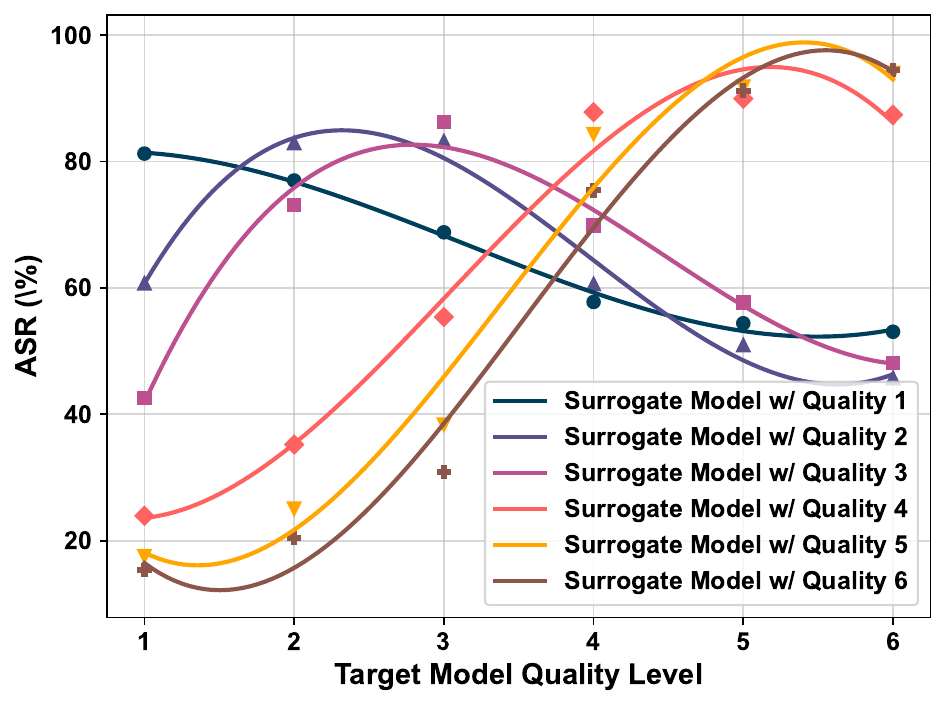} 
        \caption{ASR of PGD black-box attack results on quality level 1 to 6 of \texttt{cheng2020} \cite{cheng2020learned} model with $\epsilon=16$, $\alpha=2$, $iters=20$.}
        \label{fig:curve_bb_attack}
    \end{minipage}
    

\end{figure*}

Next, we explore the robustness of the LICCS with black-box attacks and its transferability. The concept of \textbf{``quality levels"} of LIC distinguishes LICCS from the traditional classification model, introducing unknown images to attackers and increasing the challenge of black-box attacks. Consequently, we aim to explore this unique feature by launching black-box attacks across different quality levels to evaluate the transferability between them. Table \ref{table:blackboxresults} demonstrates the top-1 accuracy of LICCS after the PGD attack~\cite{madrytowards} on \texttt{cheng2020}~\cite{cheng2020learned} and ResNet-20 on the CIFAR-10 dataset with different quality levels. The number at the beginning of the row and column denotes the quality level of the surrogate model and target model, respectively. Here, the surrogate model refers to a well-trained LIC model whose quality levels, parameters, and architecture are known to the attacker. In contrast, such information about the target models remains unknown. The experimental setup details are illustrated in Section \ref{sec:exp}. 

As demonstrated in Table \ref{table:blackboxresults}, \textbf{the attack fails to achieve effective transferability across different quality levels.} More detailed results are shown in Fig. \ref{fig:blackboxattack}. For instance, the top-1 accuracy of the target model with the quality level 6 remains robust at 87.19$\%$, indicating a highly restricted transferability from quality level 1 to 6. Such a significant discrepancy concludes that \textbf{it is challenging for naive black-box attack to defeat LICCS without improving transferability}. Consequently, as a LICCS attacker, we strive to design an attack strategy with improved transferability.

\subsection{Observation}
\label{subsec:observation}

As pointed out by several studies a natural way of improving the transferability of adversarial examples is by attacking an ensemble of models, which diversifies the surrogate models and helps to capture better the intrinsic adversarial vulnerability of the target model~\cite{tramer2018ensemble,demontis2019adversarial}.
A range of LIC models with varying quality levels could potentially be utilized as components of ensemble models. However, in LIC, the quality level is controlled through a hyper-parameter which can take arbitrary real values; yet it is infeasible to incorporate an infinite number of such models to cover all possible values due to memory and computational constraints.

To analyze the influence of the adversarial examples on each quality level, we conduct the black-box attack shown in Fig. \ref{fig:curve_bb_attack}. Findings from the figure and Table \ref{table:blackboxresults} reveal that $\texttt{ASR}(i,q) \propto 1 / \texttt{Distance(i, q)}$, where $\texttt{ASR}(i,q) = 100\% - \texttt{Acc}(i, q) $ denotes the ASR (Attack Success Rate) when attacking the surrogate quality level $i$ and evaluate on target quality level $q$. In other words, \textbf{\textit{when a specific quality level is attacked, the adjacent quality levels tend to experience more substantial impact, whereas the impact diminishes for quality levels that are more distant}}.

The observations indicate that each quality level possesses its unique sphere of influence. This implies that even if selecting the most influential model, attacking a single quality level can only affect its near vicinity and has less effect on others that are further away. Hence, it becomes crucial to identify the optimal combination of multiple quality levels rather than repeatedly attacking the individuals with the most influence.


\subsection{Saliency Score-based Sampling}
 
Building upon the observations outlined in Section \ref{subsec:observation}, we propose a method to improve the transferability of attacks across all quality levels. We first sample $\lambda_{1,2,\cdots, N}$ (the coefficient to control the rate-distortion trade-off) and train corresponding $N$ surrogate models with diverse quality levels differently. To ensure the exact black-box attack process, the $\lambda$ of training surrogate models are independent of those used in target models. Subsequently, we calculate the black-box ASR of the surrogate models and derive the $\texttt{ASR}(q_1, q_2), q_1, q_2 \in [1,N]$, to analyze the influence of each pair of quality level. Since the potential target models in real-world scenarios have an infinite range of quality levels, we strive to accurately depict the influence at each potential real number of quality levels. To achieve this, we convert the discrete ASR points into a continuous polynomial ASR curve with a polynomial fitting function $\texttt{polyn}(\cdot)$. This representation captures the ASR values across a continuous spectrum of quality levels rather than only at $N$ discrete quality levels.

To identify the optimal combination of quality levels, we introduce the coverage function, defined as $\max(A, B)$, which quantifies the collective coverage of curve $A$ and $B$. Then, we employ an accumulated integral function to calculate the saliency score for each combination of $K$ quality levels. The saliency score represents the degree to which the combination maximizes the coverage area, thereby representing the potential influence on ASR across all quality levels. Given the desired number of ensemble models $K$, where $ K < N$, the formulation of the saliency score is as follows:
\vspace{-2mm}
\begin{equation}
\vspace{-2mm}
\begin{aligned}
S(\boldsymbol{q}) = \int_{x_{min}}^{x_{max}} \max (\texttt{polyn}(q_0), \texttt{polyn}(q_1), 
\cdots, \texttt{polyn}(q_K)) \enspace dx \\
\end{aligned}
\label{eqn:method}
\end{equation}
where $\boldsymbol{q} = [ q_0, q_1, \cdots, q_K ]$ represents the combination of $K$ quality levels. $x_{min}$ and $x_{max}$ are the lowest and highest quality level of surrogate models.

After calculation, we select the largest $S$ among total $\tbinom{N}{K}$ values, corresponding to the combination of $K$ quality levels that cause the most substantial impact across the entire range of quality levels within the surrogate models.
\vspace{-2mm}
\begin{equation}
\vspace{-2mm}
\begin{aligned}
\argmax \limits_{i} \sum_{q=q^*_0}^{q^*_K} f({g_s}_{q}(Q({g_a}_{q}(\bmmc{X}+\bmmc{\delta}))))_{i} \neq y, 
\quad \textrm{s.t.} \enspace \| \bmmc{\delta} \|_{p} \leq \epsilon, \\
\end{aligned}
\label{eqn:lic_attack}
\end{equation}
where $q^*$ is from the optimal combination of quality levels corresponding to the largest $S$. A smaller value of $K$ indicates that we use fewer surrogate models to generate the transferable adversarial perturbations, thereby improving efficiency.

\section{Experiments}
\label{sec:exp}

\textbf{Setting.} We adopt PGD~\cite{madrytowards} and FGSM~\cite{43405} attack methods with \texttt{cheng2020} \cite{cheng2020learned} and \texttt{hyper}~\cite{balle2018variational} LIC models to evaluate the effectiveness of transferability across quality levels and different architectures. In this paper, we use the differential  approximation quantization from \cite{shin2017jpeg} to execute the gradient-based attack. We fix the classification model as ResNet-20. The evaluations are conducted on the CIFAR-10 dataset.

\textbf{Metric.} The attack performance is evaluated based on the top-1 accuracy of the victim LICCS. The lower accuracy of the victim model indicates better attack performance. We also measure the adversarial perturbations generation time (average of 100 times) per image.

\textbf{Hyperparameter.} For the white-box attack in Table \ref{table:whiteboxresults}, we set perturbation budget $\epsilon \in \{1, 2, 4, 8, 16\}$, learning step $\alpha\in \{1, 2\}$, and iterations $T \in \{10, 20\}$. For the black-box attack in Table \ref{table:blackboxresults}, experiments are conducted with $\epsilon \in \{4, 8\}$, $\alpha \in \{1, 2\}$, $T \in \{10, 20\}$. For the Table \ref{table:blackbox_qualityresults}, \ref{table:blackbox_hyper_qualityresults}, and \ref{table:blackboxresults}, experiments are conducted with $\epsilon \in \{4, 8, 16\}$, $\alpha \in \{1, 2\}$. All $\epsilon$ and $\alpha$ are divided by 255 to match the normalized image. The coefficients of the rate-distortion trade-off of surrogate models, $\lambda$, are randomly sampled to control the quality levels independent of those in target models. Surrogate models with $\lambda$ are fully trained following the setting of \cite{begaint2020compressai}. 

\begin{table}[ht]
\vspace{-2mm}
\setlength{\tabcolsep}{14pt}
    \begin{center}
    \caption{Top-1 accuracy of LICCS with the surrogate model \texttt{cheng2020} and target model \texttt{cheng2020} attacked by PGD. Lower accuracy demonstrates higher transferability.}
	\scalebox{0.7}{
		\begin{tabular}{c|cccccc|c|c}
			\toprule
               Quality & 1 & 2 & 3 & 4 & 5 & 6 & Average & Time \\ \hline
			\multicolumn{9}{c}{$\epsilon=4$, \quad $\alpha=1$, \quad $iters=10$} \\ 
			\midrule
                R-En & 44.32\% &  51.10\% &  54.95\% &  56.16\% &  55.10\% &  58.02\% & 53.28\% & 1.1s \\ \hline
                \rowcolor{Gainsboro!60}\textbf{Ours} & 47.86\% &  54.46\% &  51.41\% &  47.36\% &  39.26\% &  40.53\% &  \textbf{46.81\%} & 1.1s \\ \hline
			\multicolumn{9}{c}{$\epsilon=8$, \quad $\alpha=2$, \quad $iters=10$} \\ 
   			\midrule
      
                R-En & 39.83\% &  43.34\% &  45.42\% &  43.20\% &  43.23\% &  46.59\% & 43.60\% & 1.1s \\ \hline
                \rowcolor{Gainsboro!60}\textbf{Ours} & 45.52\% &  47.25\% &  42.14\% &  31.26\% &  25.01\% &  25.33\% & \textbf{36.09}\% & 1.1s \\ \hline
                
			\multicolumn{9}{c}{$\epsilon=16$, \quad $\alpha=2$, \quad $iters=10$} \\ 
   			\midrule
                R-En & 35.68\% &  37.15\% &  37.87\% &  35.76\% &  37.86\% &  40.92\% & 37.54\% & 1.1s \\ \hline
                \rowcolor{Gainsboro!60}\textbf{Ours} & 43.31\% &  41.18\% &  34.46\% &  22.64\% &  20.24\% &  20.39\% & \textbf{30.37}\% & 1.1s \\ \hline
    
			\bottomrule
		\end{tabular}
		}
	 \label{table:blackbox_qualityresults}
\end{center}
\vspace{-2mm}
\end{table}

\begin{table}[ht]
\setlength{\tabcolsep}{5pt}
    \begin{center}
    \caption {Top-1 accuracy of LICCS with the surrogate model \texttt{hyper} and target model \texttt{hyper} attacked by FGSM. Lower accuracy demonstrates higher transferability. }
	\scalebox{0.8}{
		\begin{tabular}{c|cccccccc|c|c}
 			\toprule
               Quality & 1 & 2 & 3 & 4 & 5 & 6 & 7 & 8 & Average & Time\\ \hline
			\multicolumn{11}{c}{$\epsilon=4$, \quad $iters=1$} \\ 
			\midrule
        R-En & 42.97\% & 56.03\% & 62.81\% & 67.65\% & 71.92\% & 77.36\% & 80.69\% & 83.10\% & 67.82\%  & 0.15s \\ \hline
        \rowcolor{Gainsboro!60}\textbf{Ours} & 44.69\% & 56.79\% & 61.32\% & 59.82\% & 61.75\% & 62.21\% & 67.06\% & 69.58\% & \textbf{60.28\%} & 0.15s \\ \hline
         
			\multicolumn{11}{c}{$\epsilon=8$, \quad $iters=1$} \\ 
   			\midrule
        R-En & 43.44\% & 53.26\% & 54.35\% & 56.24\% & 62.37\% & 70.22\% & 72.43\% & 72.44\% & 59.57\% & 0.15s \\ \hline
        \rowcolor{Gainsboro!60}\textbf{Ours} & 41.93\% & 47.81\% & 47.24\% & 47.89\% & 52.42\% & 55.49\% & 57.37\% & 57.38\% & \textbf{50.94\%} & 0.15s \\ \hline
         
		\end{tabular}
		}
	 \label{table:blackbox_hyper_qualityresults}
\end{center}
\vspace{-2mm}
\end{table}





\textbf{Transferablity across quality levels.} We initially evaluate the performance of our method by applying the PGD on the surrogate model \texttt{cheng2020} to generate adversarial perturbations. To evaluate the transferability across quality levels, we utilize these perturbations to disrupt a target model of unknown quality levels. The baseline method, termed ``R-En", involves randomly selecting $K$ surrogate models to conduct the ensemble attacks. To ensure consistency, we set $K=2$ for both R-En and our proposed method. To mitigate the variance arising from the randomness, we calculate the average of 20 experiment results for the R-En method. As demonstrated in Table \ref{table:blackbox_qualityresults}, compared to R-En, our method improves the average ASR by 6.46$\%$, 7.51$\%$, 7.17$\%$ with $\epsilon=4, \epsilon=8, \epsilon=16$, while maintaining the same perturbation generation time. We further measure our approach using the FGSM attack on the \texttt{hyper} model to assess the improved transferability. According to Table \ref{table:blackbox_hyper_qualityresults}, compared to R-En, our method enhances the average ASR by 7.54$\%$ and 8.63$\%$ with $\epsilon=4$ and $\epsilon=8$, respectively.

\begin{figure*}[ht]
\centering
  \includegraphics[width=1\linewidth]{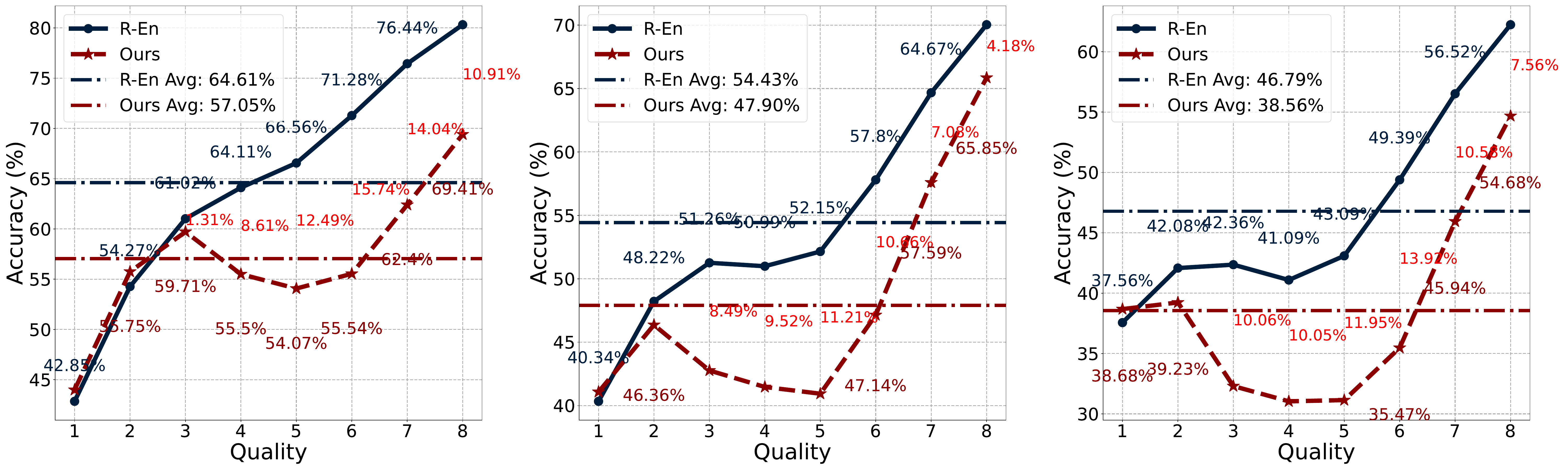}
\caption{Top-1 accuracy of LICCS with the surrogate model \texttt{cheng2020} and target model \texttt{hyper} attacked by PGD. Lower accuracy demonstrates higher transferability.}
\label{fig:transfer_arch}
\vspace{-2mm}
\end{figure*}

\textbf{Transferablity across different architectures.} To evaluate the transferability across various architectures, we apply adversarial perturbations generated from $\texttt{cheng2020}$ to disrupt an unseen target model $\texttt{hyper}$. As illustrated in Fig. \ref{fig:transfer_arch}, compared to R-En, our method can improve the average ASR by 7.57$\%$, 6.53$\%$, and 8.23$\%$ with $\epsilon=4$, $\epsilon=8$, and $\epsilon=16$, respectively.

\begin{table}[ht]
\setlength{\tabcolsep}{18pt}
    \begin{center}
        \caption {Top-1 accuracy of LICCS after PGD white-box attack on quality level 1 to 6 of \texttt{cheng2020} \cite{cheng2020learned} model. ``w/o LIC" and ``w/ LIC" mean the generated adversarial examples are fed into the ResNet-20 without LIC module and standard LICCS (ResNet-20 with LIC module), respectively. }
	\scalebox{0.7}{
		\begin{tabular}{ccccccccc}
			\toprule
			\multirow{3}{*}{\makecell{Quality \\ Level}}& $\epsilon=1$ & $\epsilon=2$ & $\epsilon=4$ & $\epsilon=8$ & $\epsilon=8$ & $\epsilon=16$ & $\epsilon=16$ \\ 
                \cline{2-8}
			& $\alpha=1$ & $\alpha=1$ & $\alpha=1$ & $\alpha=2$ & $\alpha=2$ & $\alpha=2$ & $\alpha=2$ \\ 
                   \cline{2-8}
			& $T=10$ & $T=10$ & $T=10$ & $T=10$ & $T=20$ & $T=10$ & $T=20$ \\
   
			\midrule
			\multicolumn{8}{c}{w/o LIC} \\ 
			\midrule
    1 & 92.45\% & 92.26\% & 91.75\% & 90.22\% & 89.58\% & 87.20\% & 84.9\% \\ \hline
    2 & 92.29\% & 92.02\% & 91.22\% & 89.01\% & 88.66\% & 85.81\% & 83.5\% \\ \hline
    3 & 91.93\% & 91.57\% & 90.34\% & 87.18\% & 87.17\% & 83.22\% & 80.17\% \\ \hline
    4 & 91.35\% & 90.36\% & 87.73\% & 80.09\% & 80.57\% & 73.96\% & 68.80\% \\ \hline
    5 & 90.43\% & 88.23\% & 83.16\% & 72.11\% & 71.27\% & 63.13\% & 55.24\% \\ \hline
    6 & 89.41\% & 86.05\% & 78.01\% & 63.96\% & 62.80\% & 54.57\% & 46.42\% \\
			\midrule
			\multicolumn{8}{c}{w/ LIC} \\ 
   			\midrule
    1 & 45.17\% & 41.19\% & 35.59\% & 28.45\% & 26.69\% & 22.7\% & 18.75\% \\ \hline
    2 & 53.05\% & 45.6\% & 37.86\% & 28.72\% & 25.75\% & 23.12\% & 17.07\% \\ \hline
    3 & 54.33\% & 44.47\% & 35.82\% & 25.35\% & 20.92\% & 20.41\% & 13.73\% \\ \hline
    4 & 57.24\% & 45.76\% & 37.28\% & 25.89\% & 19.31\% & 21.75\% & 12.20\% \\ \hline
    5 & 54.17\% & 42.24\% & 34.83\% & 23.06\% & 14.17\% & 19.46\% & 8.16\% \\ \hline
    6 & 50.51\% & 39.25\% & 32.53\% & 19.08\% & 9.95\% & 16.12\% & 5.53\% \\
			\bottomrule
		\end{tabular}
		}
	 \label{table:whiteboxresults}
\end{center}
\vspace{-6mm}
\end{table}

\textbf{Detailed results of white-box attack.} We provide detailed results for the white-box attack. As depicted in Table \ref{table:whiteboxresults}, we initially conducted the experiments on LICCS for quality levels ranging from 1 to 6. After attacking, LICCS can only achieve less than 50$\%$ top-1 accuracy in most cases. Further, we also perform experiments to evaluate the performance when employing these adversarial examples on the classification model without the LIC component. \textbf{Interestingly, omitting the LIC module seems to provide some defense against these adversarial examples.} For instance, when an attack is launched on LICCS with $\epsilon=16, T=20$, the LICCS can only achieve 5.53$\%$ top-1 accuracy for quality level 6. In contrast, without the LIC module, $46.42\%$ of these adversarial examples can be classified correctly. 

\begin{figure*}[t]
\centering
  \includegraphics[width=1\linewidth]{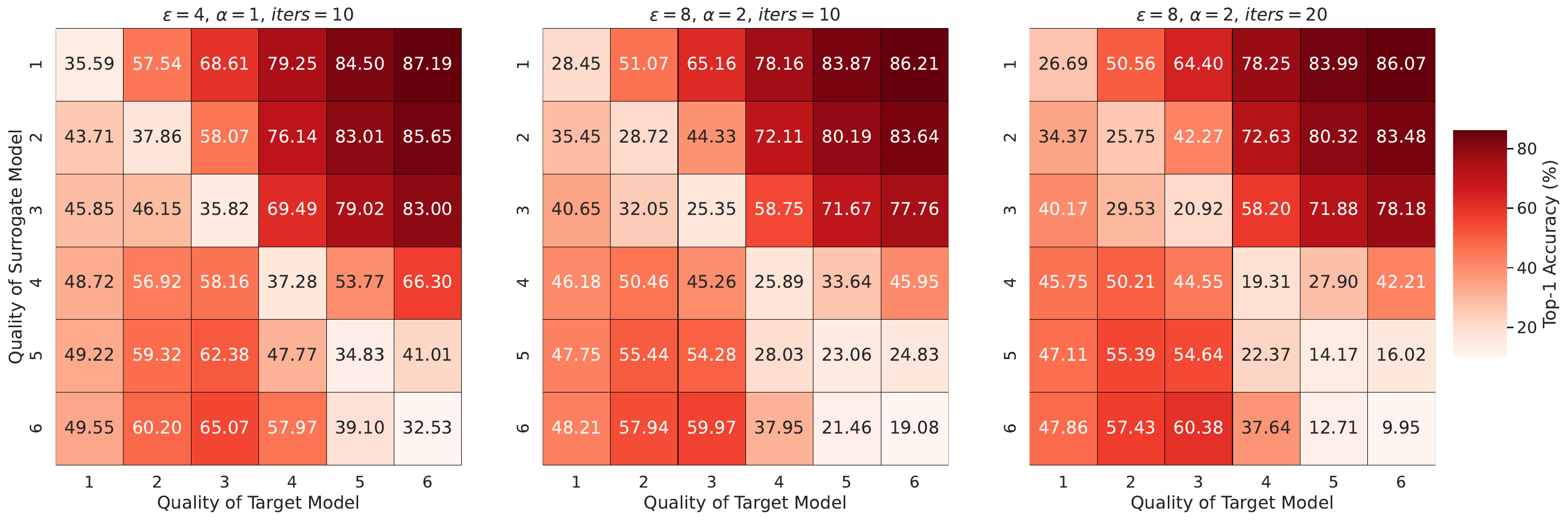}
\caption{Top-1 accuracy of LICCS after PGD black-box attack of \texttt{cheng2020} \cite{cheng2020learned} model. Each row/column corresponds to a surrogate/target model with a given quality level.}
\label{fig:blackboxattack}
\vspace{-2mm}
\end{figure*}

\textbf{Detailed results of black-box attack.} We here provide detailed black-box attacks across different quality levels to evaluate the transferability across quality levels in Fig. \ref{fig:blackboxattack} as the complementary of Table \ref{table:blackboxresults}. Each grid value represents the top-1 accuracy of the target model post-attack, using a surrogate model in a black-box setting. The experiments, as seen in Fig. \ref{fig:blackboxattack} and conducted with diverse parameters of $\epsilon, \alpha, T$, indicate varying performance when there's a mismatch or slight similarity between the surrogate and target quality levels. This underscores the limited transferability of traditional black-box attacks on LICCS across different quality levels without our method.

\section{Conclusion}
\label{sec:conclusion}
In this paper, we introduce an adversarial attack pipeline specifically designed for the LICCS and conduct a range of experiments in both white-box and black-box settings. Based on findings from the black-box experiments, we propose a saliency score-based sampling approach that allows for generating transferable perturbations even with limited model access and enhances transferability. We conducted further experiments with our methods on PGD and FGSM methods on various LIC models with multiple quality levels. The results demonstrate that our proposed method effectively enhances the transferability of adversarial perturbations across different quality levels and architectures.

\Section{References}
\bibliographystyle{IEEEbib}
\bibliography{refs}

\end{document}